\documentclass[runningheads]{llncs}
\usepackage[T1]{fontenc}
\usepackage{amsfonts,amsmath,amssymb}
\usepackage{graphicx}
\usepackage{booktabs}
\usepackage{multirow,multicol}
\usepackage{colortbl}
\usepackage{float}
\usepackage{wrapfig}
\usepackage{xcolor}     
\usepackage{pifont}
\usepackage{subcaption}
\definecolor{lightpastelpurple}{rgb}{0.69, 0.61, 0.85}
\colorlet{LightGreen}{lightpastelpurple!40}

\RequirePackage{xspace}
\makeatletter
\DeclareRobustCommand\onedot{\futurelet\@let@token\@onedot}
\def\@onedot{\ifx\@let@token.\else.\null\fi\xspace}

\def\ie{\emph{i.e}\onedot}

\makeatother
\begin{document}
\title{Symmetric masking strategy enhances the performance of Masked Image Modeling}
%
%
\author{Khanh-Binh~Nguyen\inst{1}\orcidID{0000-0002-9948-1400} \and
Chae~Jung~Park\inst{1}\orcidID{0000-0002-1261-307X}}
\authorrunning{K.B.~Nguyen et al.}
%
\institute{Division of Cancer Data Science, Cancer Big Data and Artificial Intelligence, National Cancer Center, South Korea\\
\email{\{binhnk,cjp\}@ncc.re.kr}}
\maketitle              
\begin{abstract}
Masked Image Modeling (MIM) is a technique in self-supervised learning that focuses on acquiring detailed visual representations from unlabeled images by estimating the missing pixels in randomly masked sections.
It has proven to be a powerful tool for the preliminary training of Vision Transformers (ViTs), yielding impressive results across various tasks.
Nevertheless, most MIM methods heavily depend on the random masking strategy to formulate the pretext task.
This strategy necessitates numerous trials to ascertain the optimal dropping ratio, which can be resource-intensive, requiring the model to be pre-trained for anywhere between 800 to 1600 epochs.
Furthermore, this approach may not be suitable for all datasets.
In this work, we propose a new masking strategy that effectively helps the model capture global and local features.
Based on this masking strategy, SymMIM, our proposed training pipeline for MIM is introduced.
SymMIM achieves a new SOTA accuracy of 85.9\% on ImageNet using ViT-Large and surpasses previous SOTA across downstream tasks such as image classification, semantic segmentation, object detection, instance segmentation tasks, and so on.
\end{abstract}
\section{Introduction}
\label{sec:intro}
Masked image modeling (MIM) \cite{bao2021beit,he2017mask,chen2023context,xie2022simmim} is a self-supervised learning technique that aims to learn rich visual representations from unlabeled images by predicting the missing pixels in randomly masked regions.
Inspired by masked language modeling (MLM) \cite{devlin2018bert} in natural language processing (NLP) tasks, MIM \cite{bao2021beit} has gained popularity in the context of self-supervised learning using Vision Transformers (ViTs) \cite{dosovitskiy2020image}.
In MIM, a subset of vision tokens is randomly masked, and the model learns to recover the original image through a Transformers network during training.

Several concurrent research efforts \cite{dong2023peco,li2022architecture,wei2022masked} have focused on designing patch-level dictionaries and image tokenizers to establish appropriate learning objectives (e.g., vision token IDs) for MIM.
While advanced results can be obtained, existing off-the-shelf image tokenizers (such as discrete VAE \cite{dalal2005histograms} used in BEiT) rely on additional training stages and domain-specific knowledge, resulting in a less flexible two-stage pre-training approach.

Despite this limitation, MIM methods have proven effective for pre-training vision transformers, leading to state-of-the-art performance on various downstream tasks, including image segmentation \cite{kirillov2023segment} and exploration tasks \cite{li2022exploring}.
However, it is worth noting that most MIM approaches still rely on a random masking strategy with a high masking ratio (e.g., 75\% to 95\%), which may not fully capture both the global and local features of the images.
Additionally, the computational cost associated with finding the optimal masking ratio through extensive trials remains a challenge, especially when dealing with diverse datasets.

In this work, we revisit the masking strategy employed in conventional Masked Image Modeling (MIM) and related approaches.
While conventional MIM and its derivatives are tailored for masked image modeling, they inherently encourage the network to capture fine-grained visual context and semantics.
However, we take a different approach regarding the learning objective.
Instead of using just pixel-wise MSE loss, we divide the training to two different patches with large and small size of the mask.
Then, enforcing the contrastive loss between them to encourage the consistency between global and local features.
This approach effectively structures the visual space, yielding semantically meaningful representations.

Our proposed solution introduces a symmetric masking strategy that addresses existing limitations and enhances MIM performance.
Specifically, we symmetrically mask image patches along both the horizontal and vertical axes, creating four quadrants of visible and masked patches.
This design ensures that each masked patch corresponds to a visible patch containing similar semantic and structural information, facilitating the reconstruction task and feature learning.
Moreover, our symmetric masking strategy simplifies hyperparameter tuning, relying solely on the size of the mask, thereby reducing overall complexity.

Leveraging this symmetric masking strategy, we introduce SymMIM, a straightforward yet highly effective framework for Masked Image Modeling (MIM).
SymMIM comprises a ViT encoder and projector that map the feature maps into low-level embedding space, then leveraging the contrastive learning with the EMA model to encourage the consistency.
What sets SymMIM apart is its simplicity—it avoids complex designs such as masking strategy probing, discrete VAE, or clustering.

Despite its straightforward approach, SymMIM outperforms previous MIM methods across various tasks, including image classification, object detection, and instance segmentation.
Extensive experiments conducted on the ImageNet-1K dataset demonstrate the scalability and effectiveness of SymMIM, achieving a new state-of-the-art performance on several vision benchmarks when applied to larger models and diverse datasets.

\section{Related Work}
\subsection{Self-supervised learning}
In self-supervised representation learning (SSL), a model trains on a pretext task where supervision is derived solely from the input data, without relying on labeled data.
Contrastive learning stands out as a popular approach within SSL, aiming to learn representations by contrasting positive and negative samples.
Notable methods in this category include SimCLR \cite{chen2020simple}, CPC \cite{oord2018representation}, MoCo \cite{he2020momentum}, MoCo-v2 \cite{chen2020improved}, BYOL \cite{grill2020bootstrap}, and DINO \cite{caron2021emerging}. 
Furthermore, group instance contrastive learning approaches, such as DeepCluster \cite{caron2018deep} and SwAV \cite{caron2019unsupervised}, incorporate clustering into contrastive learning to enhance the quality of learned representations.
An alternative avenue for self-supervised learning involves generative modeling, where the focus lies in acquiring a generative model capable of capturing the underlying data distribution.

\subsection{Masked Modeling}
Masked modeling learns representations by reconstructing a masked portion of the input.
Pioneering works in natural language processing (NLP) have introduced various pretraining objectives.
BERT \cite{devlin2018bert} utilizes a bidirectional transformer and demonstrates few-shot learning capabilities through masked language modeling.
GPT \cite{brown2020language,radford2019language}, on the other hand, employs autoregressive, causal masking, showcasing multitasking, few-shot, and in-context learning capabilities.
In the realm of computer vision, early works such as Stacked Denoising Autoencoders \cite{vincent2010stacked} and Context Encoder \cite{chen2023context} explored masked image modeling as a means of denoising or representation learning.

Recently, with the widespread adoption of transformers \cite{dosovitskiy2020image} as backbone architectures for vision tasks—where images are divided into patches and tokenized as sequences—researchers aim to transfer the success of language sequence modeling to large-scale vision transformers.
Notable early works, including BEiT \cite{bao2021beit}, MAE \cite{he2022masked}, and SimMIM \cite{xie2022simmim}, have explored BERT-style pretraining for vision transformers.
Interestingly, both MAE and SimMIM emphasize the necessity of a significantly higher mask ratio compared to NLP approaches to achieve effective visual representation learning.
Furthermore, recent research extends masked pretraining to hierarchical architectures \cite{chen2023mixed,xie2022simmim} and investigates the impact of data augmentation \cite{chen2023mixed,fang2022corrupted}.

CrossMAE, instead of relying on self-attention, employs cross-attention blocks in the decoder to reconstruct only a subset of the masked tokens during pretraining \cite{bachmann2022multimae,geng2022multimodal,li2023scaling}.
This approach aims to learn both effective representations and reconstruction capabilities.
In contrast, BERT-style pretraining heavily utilizes self-attention, resulting in computational complexity that scales polynomially with sequence length.

Siamese MAE \cite{gupta2023siamese}, on the other hand, adopts an asymmetric masking pattern and decodes video frames conditioned on earlier frames.
In this setting, all masked patches are reconstructed.

ConMIM \cite{yi2022masked} also leverage the contrastive loss to boost the representation learning of the MIM.
However, in contrast with ours, they propose a different type of objective base on the InfoNCE loss for each pixels while we using the contrastive loss for low-level embeddings.

This paper investigates the necessity and impact of different masking methods for pretraining feature extractors.
We conduct a comparative analysis between symmetric and random masking, considering various masking ratios, and evaluating their effects on downstream tasks.
Additionally, we propose a straightforward yet effective approach to reduce hyperparameter tuning when selecting the optimal masking ratio.

\section{Symmetric Masked Image Modeling}
\begin{figure}[!ht]
  \centering
  \includegraphics[width=\linewidth]{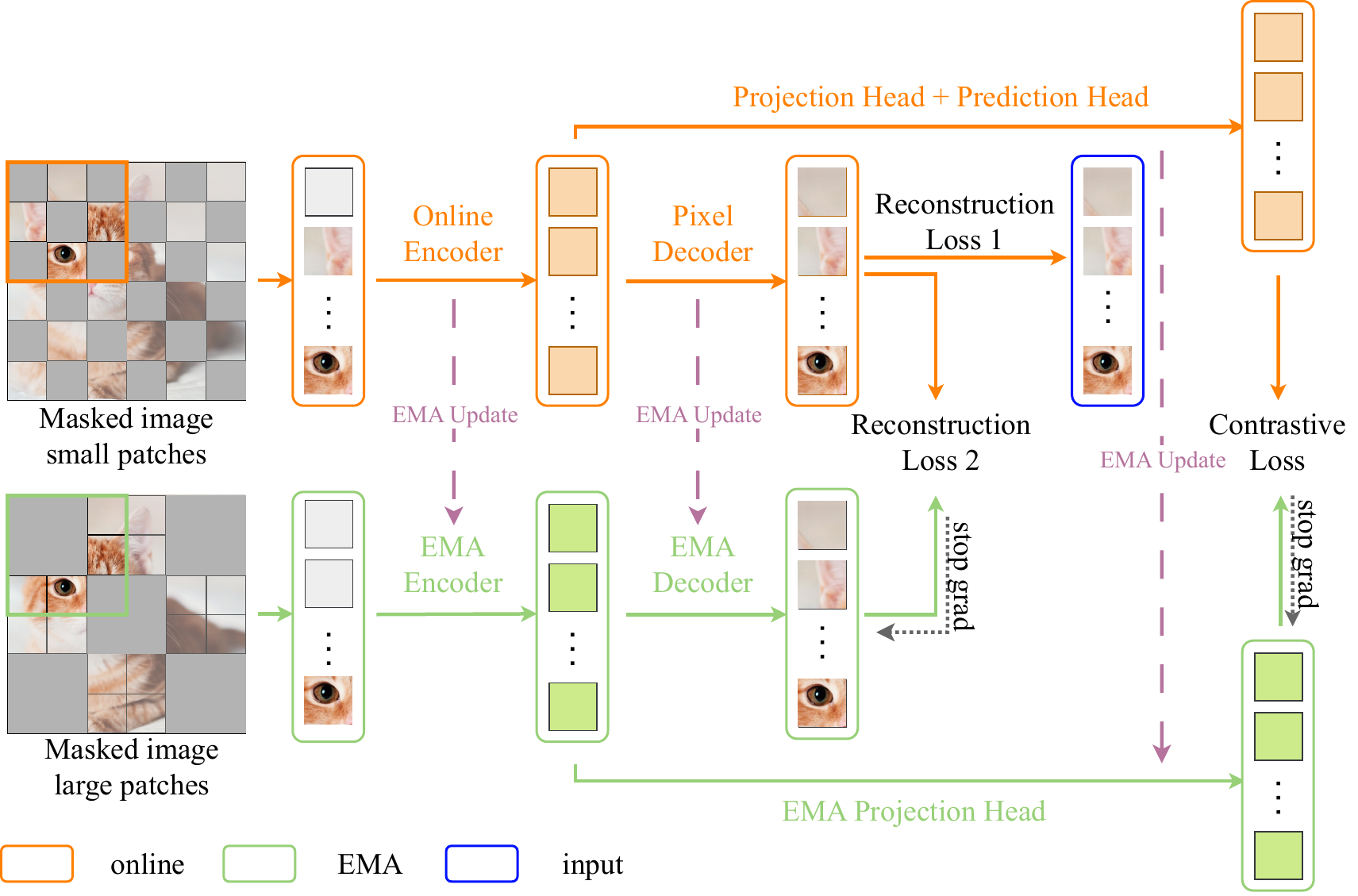}
  \caption{The pipeline of SymMIM.
  SymMIM performs the symmetric masked patch prediction and the EMA update policy from MoCo-v3 \cite{chenempirical} to effectively guides the representation learning of the trained model and enhances the network’s capability to capture more ﬁne-grained visual context.
  Following MoCo-v3 design, there is a projection head and prediction head attached to the online encoder while for the EMA encoder, there is only an EMA projection head.
  }
  \label{fig:pipeline}
\end{figure}

In this section, we delve into the details of our approach. We commence by reviewing the standard Masked Image Modeling (MIM) method and highlighting its limitations.
Subsequently, we introduce the symmetric masking strategy, which enhances the quality and diversity of masked regions. 
Building upon this strategy, we propose the Symmetric Masked Image Modeling (SymMIM) framework, which employs an asymmetric momentum contrastive learning scheme.
This scheme enables us to fully leverage the advantages of symmetric masking while achieving higher efficiency.
We substantiate the effectiveness of our approach in the subsequent sections.

\subsection{Preliminaries}
Transformer architectures \cite{dosovitskiy2020image} have gained widespread adoption in computer vision tasks.
One prominent method for visual representation learning involves a pretext task known as masked image modeling.
Specifically, a large proportion of image patches are randomly masked.
Hence, the backbone network is trained to recover the token IDs of these corrupted images.
To achieve this, a patch-level vision dictionary is employed.
This dictionary is more fine-grained and is typically static and pre-defined by an existing image tokenizer \cite{ramesh2021zero,esser2021taming}.
The image tokenizer converts continuous visual signals into discrete keys.
Notably, in the influential work called BEiT \cite{bao2021beit}, a pre-trained discrete VAE (Variational Autoencoder) serves as the tokenizer.
The ultimate goal is to predict the masked patches, which is formulated as a classification task with cross-entropy loss.
In summary, masked image modeling within transformer architectures enhances visual representation learning by leveraging masked patches and a carefully designed vision dictionary as:
\begin{equation}
    \mathcal{L}_{\text{mim}}(x)=\mathbb{E}_{j \in \mathcal{M}}\left[-\log p\left(y_j \mid f(\hat{x})_j\right)\right]
\end{equation}
where $\mathcal{M}$ denotes the set of masked patch indices, $\hat{x}$ is the corrupted image after random masking, $y_j$ is the positive key index in the patch-level dictionary, and $p(\cdot|\cdot)$ indicates the probability that correctly identifies the recovered patch $f(\hat{x})_j$ with a patch index of $j$.

\subsection{Symmetric masking strategy}
Patch alignment is a technique that divides an image into fixed-size patches, treating each patch as a token for vision Transformers.
This approach facilitates representation learning by allowing the Transformer to capture both global and local dependencies among patches, thereby learning rich semantic features from the image.
Additionally, patch alignment simplifies the masking strategy for self-supervised learning: each patch can be either fully visible or fully masked.
The Transformer learns to create or infer missing signals from the context by predicting the masked patches.

In their work, \cite{xie2022simmim} explore various patch-alignment masking strategies, including random masking, square masking, block-wise masking, and central region masking.
However, adopting these strategies often involves multiple trials to determine the optimal masking ratio for each strategy and dataset, which is not an ideal approach.
For instance,\cite{xie2022simmim} perform \textbf{more than 30 trials} to identify the optimal strategy and ratio during \textbf{pretraining with 800 epochs} and \textbf{fine-tuning for 100 epochs}.

On another note, natural images pose a challenge due to their strong spatial correlation. Pixels close to each other tend to exhibit similar values or properties.
Consequently, feature maps generated by convolutional neural networks (CNNs) may also exhibit high spatial correlation within each channel, as they derive from local neighborhoods of input pixels.
When applying dropout regularization to feature maps, some information from masked-out pixels may still be preserved by neighboring pixels. This preservation can diminish the effectiveness of the masking strategy.
During both forward and backward passes, this information propagates through the network, potentially compromising robustness against overfitting.

In order to enhance the backbone network during pre-training and make the task more challenging, we adopt the approach proposed by MAE \cite{he2022masked}, which involves masking a substantial number of patches.
These patches, such as $16\times16$, contain low information content and exhibit high semantic redundancy.
Drawing inspiration from spatial autocorrelation patterns \cite{nguyen2022checkerboard}, we introduce the symmetric masking strategy.
Specifically, instead of random image masking, we apply a checkerboard pattern to mask the image.
Leveraging this pattern allows us to flexibly control the patch size and regulate the retained information within the masked image, enabling us to focus on either local or global features.
Moreover, since the checkerboard pattern maintains a fixed masking ratio (approximately 50\%), we eliminate the need to search for an optimal masking ratio.
This reduction in effort results in cost savings during both pretraining and fine-tuning across various datasets.

\subsection{Momentum contrastive learning}
In this paper, we propose a masking strategy that dynamically adjusts the size of the checkerboard pattern to strike a balance between local and global information within the input image.
Specifically, we employ different mask sizes for the online network and the momentum encoder—two models sharing the same backbone but with distinct update rates.

\begin{enumerate}
    \item The \textit{online network} employs smaller masks, effectively concealing more local information. This deliberate difficulty enhances the learning task.
    \item The \textit{momentum encoder}, on the other hand, utilizes larger masks, effectively hiding more global information. This approach promotes consistency and diversity in the learned key features.
    \item The \textit{momentum encoder parameters} are updated using a moving average of the \textit{online network parameters}.
\end{enumerate}

The reconstruction loss is computed in two ways:
\begin{itemize}
    \item Between the original image and the recovered image from the online network.
    \item Between the recovered images from both the online network and the momentum encoder.
\end{itemize}

The loss function is defined as follows:

\begin{equation}
\begin{split}
\label{eq:mim}
\mathcal{L}_{\text{mim}}(x) &= \mathcal{L}_{\text{rec}_1} + \mathcal{L}_{\text{rec}_2}\\ 
&= \mathbb{E}_{j \in \mathcal{M}_1}\left[-\log p\left(y_j \mid f(\hat{x})_j\right)\right] \\
&+ \mathbb{E}_{j \in \mathcal{M}_1 \cap \mathcal{M}_2}\left[-\log p\left(\tilde{f}(\hat{x})_j \mid f(\hat{x})_j\right)\right]
\end{split}
\end{equation}

where $f(\hat{x})_j$ and $\tilde{f}(\hat{x})_j$ represent the patches recovered by the online network and the momentum network, respectively, with patch index $j$.
$M_1$ and $M_2$ denote the sets of masked patch indices from the online network and the momentum encoder, respectively.

After the encoder, we apply a projection head and a prediction head to transform feature maps into an embedding representation.
Notably, in the momentum path, only a projection head is applied, following the MoCo-v3 approach to ensure asymmetry and prevent collapse during training.

An InfoNCE loss \cite{van2018inaturalist} is employed to regularize this representation.
The goal is to bring it closer to its positive key $k+$ while distancing it from negative keys.
The loss function is defined as:

\begin{equation}
\label{eq:con}
    \mathcal{L}_{\text{con}}(x) = \mathbb{E}_{j \in \mathcal{M}_1}\left[-\log \frac{\exp \left(\left\langle g(\hat{x})_j, \operatorname{sg}\left[\tilde{g}(\hat{x})_j\right]\right\rangle / \tau\right)}{\sum_{i=1}^K \exp \left(\left\langle g(\hat{x})_j, \operatorname{sg}\left[\tilde{g}(\hat{x})_i\right]\right\rangle / \tau\right)}\right]
\end{equation}
where $\left\langle g(\hat{x}), \tilde{g}(\hat{x})\right\rangle$ represents the cosine similarity measured by the dot product between two $L_2$-normalized features: $g(\hat{x})$ and $\tilde{g}(\hat{x})$ from the projectors.
$\tau$ is the temperature hyperparameter.
$k$ is the dynamic key.
$K$ is the dictionary size.
$\operatorname{sg}\left[\cdot\right]$ indicates the stop-gradient operation.

The positive key corresponds to the features generated from the momentum network.
The overall training objective is defined as:

\begin{equation}
\label{eq:mimcon}
    \mathcal{L} = \mathcal{L}_{\text{mim}} + \lambda \mathcal{L}_{\text{con}}
\end{equation}

Here, $\lambda$ controls the weight of the loss (for simplicity, it is set to 1).
With these training objectives, the backbone network is motivated to better capture visual context and learn to encode both local and global discriminative representations.

The parameters of the momentum encoder and momentum projector are updated using the exponential moving average strategy:

\begin{equation}
    \theta_k \leftarrow m \theta_k + (1 - m) \theta_q
\end{equation}
where $m \in [0,1)$ represents the momentum coefficient.

\section{Experiments}
\label{sec:experiments}
We conduct self-supervised pretraining on the ImageNet-1K dataset, following the methodologies of MAE \cite{he2022masked} and SimMIM \cite{xie2022simmim}. The hyperparameters align with those used in SimMIM for both the ViT-Base and ViT-Large backbones. Specifically:
\begin{itemize}
    \item We employ a linear prediction head with a target image size of $224\times224$, utilizing an $L_1$ loss for masked pixel prediction.
    \item The projection head consists of a 3-layer MLP, following the approach in \cite{chen2020simple}.
    \item The hidden layers of both MLPs have a dimensionality of 4096 with ReLU activation, while the output layers are 256-dimensional without ReLU.
    \item We set the temperature parameter $\tau$ to $0.1$ by default.
    \item The large-size mask is constructed as the default configuration, with a patch size of $32\times32$, while the small-size mask has a patch size of $16\times16$. This design ensures a trade-off between local and global information without excessively masking data.
\end{itemize}

\subsection{ImageNet Classification}
We fine-tuning pre-trained models on the ImageNet-1K dataset \cite{deng2009imagenet} to benchmark our proposed SymMIM.
Our fine-tuning setup closely follows that of BEiT \cite{bao2021beit}.
Specifically:
\begin{itemize}
    \item For ViT-Small, fine-tuning for total 200 epochs with a warm-up of 5 epochs and a layer decay of 0.8.
    \item For ViT-Base fine-tuning for total 100 epochs with a warm-up of 20 epochs and a layer decay of 0.65.
    \item For ViT-Large fine-tuning for total 50 epochs with a warm-up of 5 epochs and a layer decay of 0.75.
\end{itemize}

For fair comparison, we only compare with previous SOTA that fine-tuning on the same input image size as $224\times224$ pixels.
Our observations reveal that models with the proposed symmetric masking strategy consistently surpass previous approaches.
Comparing with previous SOTA such as SimMIM \cite{xie2022simmim}, ConMIM \cite{yi2022masked}, MAE \cite{he2022masked}, SymMIM outperforms with a significant margin.
However, SymMIM lacks behind the ConMIM$^{\dagger}$ \cite{yi2022masked} which fine-tuning with higher resolution of $384\times384$.

\begin{table}[!ht]
  \caption{ImageNet-1K Classification Accuracy Comparison.
    In this comparison, SymMIM demonstrates performance that is on par with or even superior to other state-of-the-art self-supervised learning methods.
    All experiments were conducted over 800 epochs.
    Notably, the best results are highlighted in bold, while the second best results are underlined.
    Additionally, the symbol $^{\dagger}$ denotes that a resolution of  $384\times384$ was used for fine-tuning
  }
  \label{tab:vitb}
  \centering
  \begin{tabular}{@{}lllll@{}}
    \toprule
    Method                                          &   Pre-train data    &   ViT-S   &   ViT-B   &   ViT-L\\
    \midrule
    Supervised \cite{he2022masked,steiner2021train} &   -           &   79.0    &   82.3    &   82.6\\
    DINO \cite{caron2021emerging}                   &   IN1K        &   -       &   82.8    &   - \\
    MoCo-v3 \cite{chenempirical}                    &   IN1K        &   81.4    &   83.2    &   84.1\\
    BEiT \cite{bao2021beit}                         &   IN1K + DALLE&   -       &   83.2    &   85.2\\
    MultiMAE \cite{bachmann2022multimae}            &   IN1K        &   -       &   83.3    &   -\\
    MixedAE \cite{chen2023mixed}                    &   IN1K        &   -       &   83.5    &   -\\
    CIM \cite{fang2022corrupted}                    &   IN1K        &   81.6    &   83.3    &   -\\
    MAE \cite{he2022masked}                         &   IN1K        &   78.9    &   83.3    &   85.4\\
    SimMIM \cite{xie2022simmim}                     &   IN1K        &   -       &   83.8    &   -\\
    ConMIM \cite{yi2022masked}                      &   IN1K        &   \underline{82.0}    &   \underline{83.7}    &   \underline{85.5}\\
    CrossMAE \cite{fu2024rethinking}                &   IN1K        &   79.3    &   83.7    &   85.4 \\
    ConMIM$^{\dagger}$ \cite{yi2022masked}          &   IN1K        &   83.9    &   85.3    &   86.5\\
    \midrule
    \rowcolor{LightGreen}SymMIM (Ours)              &   IN1K        &   \textbf{83.0}    &   \textbf{84.0}    &   \textbf{85.7}\\
  \bottomrule
  \end{tabular}
\end{table}

For further evaluation, we record the results for ViT-Small and ViT-Large in \ref{tab:vits} and \ref{tab:vitl}, respectively.
It clearly shows that SymMIM outperforms previous SOTA methods when using ViT-Small and ViT-Base as a backbone and is on par with MAE \cite{he2022masked} when using ViT-Large.
Notably, although ConMIM \cite{yi2022masked} achieves higher performance compared with our proposed SymMIM, they fine-tune the model on a larger resolution input of $384\times384$.

\begin{table}[!ht]
    \caption{Top-1 Accuracy (\%) for ImageNet-1K \cite{deng2009imagenet} Classification with ViT-Small \cite{dosovitskiy2020image}.
    In this comparison, various models utilize ImageNet-1K data for pre-training, employing a resolution of $224\times224$. Notably, for fine-tuning, a higher resolution of $384\times384$ is indicated by the symbol $^{\dagger}$.
    }
    \label{tab:vits}
    \centering
    \begin{tabular}{@{}lccc@{}} 
    \toprule
        Method & Arch. & \#Epochs & Acc. \\
        \midrule \textcolor{gray}{scratch} & \textcolor{gray}{ViT-S/16} & \textcolor{gray}{-} & \textcolor{gray}{79.8} \\
        MoCo-v3 \cite{chenempirical} & ViT-S/16 & 600 & 81.4 \\
        DINO \cite{caron2021emerging} & ViT-S/16 & 1600 & 81.5 \\
        \midrule BEiT \cite{bao2021beit} & ViT-S/16 & 300 & 81.3 \\
        ConMIM \cite{yi2022masked} & ViT-S/16 & 300 & 82.0 \\
        iBOT \cite{zhou2021ibot} & ViT-S/16 & 600 & 81.5 \\
        SimMIM \cite{xie2022simmim} & ViT-S/16 & 800 & 80.1 \\
        MAE \cite{he2022masked} & ViT-S/16 & 800 & 80.9 \\
        iBOT \cite{zhou2021ibot} & ViT-S/16 & 3200 & 82.3 \\
        \rowcolor{LightGreen}SymMIM & ViT-S/16 & 300 & \textbf{83.0}\\
        \midrule
        ConMIM$^{\dagger}$ \cite{yi2022masked} & ViT-S/16 & 300 & 83.9\\
    \bottomrule
    \end{tabular}
\end{table}

\begin{table}[!ht]
    \caption{Top-1 Accuracy (\%) for ImageNet-1K \cite{deng2009imagenet} Classification with ViT-Large \cite{dosovitskiy2020image}.
    In this comparison, various models utilize ImageNet-1K data for pre-training, employing a resolution of $224\times224$. Notably, for fine-tuning, a higher resolution of $384\times384$ is indicated by the symbol $^{\dagger}$
    }
    \label{tab:vitl}
    \centering
    \begin{tabular}{@{}lccc@{}} 
    \toprule
        Method & Arch. & \#Epochs & Acc. \\
        \midrule \textcolor{gray}{scratch} & \textcolor{gray}{ViT-L/16} & \textcolor{gray}{-} & \textcolor{gray}{82.6} \\
        MoCo-v3 \cite{chenempirical} & ViT-L/16 & 600 & 84.1 \\
        \midrule MAE \cite{he2022masked} & ViT-L/16 & 400 & 84.3 \\
        ConMIM \cite{yi2022masked} & ViT-L/16 & 400 & 84.6 \\
        BEiT \cite{bao2021beit} & ViT-L/16 & 800 & 85.2 \\
        MAE \cite{he2022masked} & ViT-L/16 & 800 & 85.2 \\
        ConMIM \cite{yi2022masked} & ViT-L/16 & 800 & 85.2 \\
        iBOT \cite{zhou2021ibot} & ViT-L/16 & 1000 & 84.8 \\
        ConMIM \cite{yi2022masked} & ViT-L/16 & 1600 & 85.5 \\
        MAE \cite{he2022masked} & ViT-L/16 & 1600 & 85.9 \\
        \rowcolor{LightGreen}SymMIM & ViT-L/16 & 1600 & \textbf{85.9}\\
        \midrule BEiT$^{\dagger}$ \cite{bao2021beit} & ViT-L/16 & 800 & 86.3 \\
        ConMIM$^{\dagger}$ \cite{yi2022masked} & ViT-L/16 & 800 & 86.3 \\
        ConMIM$^{\dagger}$ \cite{yi2022masked} & ViT-L/16 & 1600 & 86.5 \\
    \bottomrule
    \end{tabular}
\end{table}

\subsection{Object Detection and Instance Segmentation}
For object detection and instance segmentation, we conduct an end-to-end fine-tuning of Mask R-CNN \cite{he2017mask} on the COCO dataset \cite{lin2014microsoft}.
Our approach involves replacing most attention blocks in the ViT backbone with windowed blocks to mitigate the quadratic complexity associated with self-attention.
However, we retain four global blocks for cross-window interaction.
Additionally, we evenly integrate four up-and-down-sample modules with the ViT architecture to generate the pyramid features required by FPN.

Our training setup closely follows the implementation \cite{li2021benchmarking}.
Specifically, we resize input images to $1024 \times 1024$ and augment them using large-scale jitter \cite{ghiasi2021simple} with a resize scale ranging from 0.1 to 2.0.
The pre-trained models undergo 25 epochs of fine-tuning using the AdamW optimizer \cite{loshchilov2017decoupled} with a weight decay of 0.1 and a cosine learning rate scheduler.

To ensure fair comparisons, we specifically tune the learning rate for each model, following the guidelines in \cite{he2022masked}.
Remarkably, our approach outperforms SimMIM with fewer pre-training epochs and achieves comparable improvements to SymMIM using the same pre-trained epochs on ViT-Small.
Similar enhancements are observed for ViT-Base. Detailed results can be found in \ref{tab:coco}.

\begin{table}[!ht]
    \caption{Object detection and instance segmentation on COCO \cite{lin2014microsoft} in terms of $\mathrm{AP}^{\text{box}}(\%)$ and $\mathrm{AP}^{\text{mask}}(\%)$.
    We tune the optimal learning rate for each model \cite{xie2022simmim}.
    symbol $^{\dagger}$ indicates transferring after intermediate fine-tuning on ImageNet-1K \cite{deng2009imagenet}, which is a common practice of BERT \cite{devlin2018bert}.
    }
    \label{tab:coco}
    \centering
    \begin{tabular}{@{}l|c|c|c|c@{}}
        \toprule
             Methods & Arch. & \#Ep. & $\mathrm{AP}^{\text {box }}$ & $\mathrm{AP}^{\text {mask }}$ \\
             \midrule
             \textcolor{gray}{scratch} & \textcolor{gray}{ViT-S/16} & \textcolor{gray}{-} & \textcolor{gray}{43.1} & \textcolor{gray}{38.8} \\
             MAE \cite{he2022masked} & ViT-S/16 & 800 & 38.9 & 35.6 \\
             ConMIM$^{\dagger}$ \cite{yi2022masked} & ViT-S/16 & 300 & 45.8 & 41.0 \\
             \rowcolor{LightGreen} SymMIM$^{\dagger}$ \cite{xie2022simmim} & ViT-S/16 & 300 & \textbf{46.0} & \textbf{41.7} \\
             MAE$^{\dagger}$ \cite{he2022masked} & ViT-S/16 & 800 & 41.5 & 37.8 \\
             SimMIM$^{\dagger}$ \cite{xie2022simmim} & ViT-S/16 & 800 & 43.0 & 38.7 \\
             \midrule
             \textcolor{gray}{scratch} & \textcolor{gray}{ViT-B/16} & \textcolor{gray}{-} & \textcolor{gray}{46.5} & \textcolor{gray}{41.7} \\
             MoCo-v3 \cite{chenempirical} & ViT-B/16 & 600 & 47.3 & 42.2 \\
             BEiT \cite{bao2021beit} & ViT-B/16 & 800 & 47.4 & 42.1 \\
             ConMIM \cite{yi2022masked} & ViT-B/16 & 800 & 47.8 & 42.5 \\
             \rowcolor{LightGreen} SymMIM & ViT-B/16 & 800 & 48.7 & 43.3 \\
             SimMIM \cite{xie2022simmim} & ViT-B/16 & 800 & 48.7 & 43.2 \\
             DINO \cite{caron2021emerging} & ViT-B/16 & 1600 & 47.6 & 42.3 \\
             iBOT \cite{zhou2021ibot} & ViT-B/16 & 1600 & 48.3 & 42.7 \\
             MAE \cite{he2022masked} & ViT-B/16 & 1600 & 48.0 & 43.0 \\
             BEiT$^{\dagger}$ \cite{bao2021beit} & ViT-B/16 & 800 & 48.2 & 43.3 \\
             SimMIM$^{\dagger}$ \cite{xie2022simmim} & ViT-B/16 & 800 & 48.4 & 43.5 \\
             ConMIM$^{\dagger}$ \cite{yi2022masked} & ViT-B/16 & 800 & 48.7 & 43.6 \\
             MAE$^{\dagger}$ \cite{he2022masked} & ViT-B/16 & 1600 & 47.8 & 42.9 \\
             \rowcolor{LightGreen} SymMIM$^{\dagger}$ & ViT-B/16 & 800 & \textbf{48.8} & \textbf{43.7} \\
        \bottomrule
    \end{tabular}
\end{table}

\subsection{Semantic Segmentation}
We assess the performance of SymMIM in downstream semantic segmentation using the ADE20K \cite{zhou2017scene} benchmark, which comprises 25,000 images across 150 semantic categories.
Our evaluation metric is the mean intersection over union (mIOU).
For this task, we employ UperNet \cite{xiao2018unified} and follow the same setup as BEiT \cite{bao2021beit}.
The input images are resized to $512\times512$, and the model undergoes 160,000 iterations of fine-tuning.
Additionally, we leverage intermediate fine-tuning to fully exploit the potential of pre-trained models, as per BEiT's approach \cite{bao2021beit}.
Specifically, we first fine-tune the models on ImageNet-1K classification and then transfer them to the ADE20K dataset.
Our results, detailed in \ref{tab:pascal}, consistently outperform the baseline BEiT, achieving significant improvements.
For instance, our approach achieves 50.8\% mIOU compared to BEiT's 47.7\% on ViT-Base.
Remarkably, even our SymMIM using ViT-Small achieves comparable performance to BEiT using ViT-Base (\ie 47.7\%).
Notably, models pre-trained via masked image modeling tend to exhibit better segmentation performance than those pre-trained using conventional contrastive learning methods, such as MoCo-v3 \cite{chenempirical}.
By using pure pixel-level objectives, models can acquire more complex and detailed representations at the region or pixel-level, as these results indicate.

\begin{table}[!ht]
    \caption{Semantic segmentation on ADE20K \cite{zhou2017scene} in terms of mIOU (\%).
    "\#Ep." means the number of pre-trained epochs.
   Symbol $^{\dagger}$ indicates transferring after intermediate fine-tuning on ImageNet-1K \cite{deng2009imagenet}, which is a common practice of BERT \cite{devlin2018bert}.
    }
    \label{tab:pascal}
    \centering
    \begin{tabular}{@{}l|c|c|c@{}}
    \toprule
         Methods & Arch. & \#Ep. & $\mathrm{mIOU}$ \\
         \midrule
         BEiT$^{\dagger}$ \cite{bao2021beit} & ViT-S/16 & 300 & 46.3 \\
         ConMIM$^{\dagger}$ \cite{yi2022masked} & ViT-S/16 & 300 & 47.7 \\
         \rowcolor{LightGreen} SymMIM$^{\dagger}$ & ViT-S/16 & 300 & \textbf{47.9} \\
         \midrule
         \textcolor{gray}{scratch} & \textcolor{gray}{ViT-B/16} & \textcolor{gray}{-} & \textcolor{gray}{45.3} \\
         MoCo-v3 \cite{chenempirical} & ViT-B/16 & 600 & 47.2 \\
         BEiT \cite{bao2021beit} & ViT-B/16 & 800 & 45.6 \\
         ConMIM \cite{yi2022masked} & ViT-B/16 & 800 & 46.0 \\
         DINO \cite{caron2021emerging} & ViT-B/16 & 1600 & 46.8 \\
         MAE \cite{he2022masked} & ViT-B/16 & 1600 & 48.1 \\
         iBOT \cite{zhou2021ibot} & ViT-B/16 & 1600 & 50.0 \\
         BEiT$^{\dagger}$ \cite{bao2021beit} & ViT-B/16 & 800 & 47.7 \\
         ConMIM$^{\dagger}$ \cite{yi2022masked} & ViT-B/16 & 800 & 49.8 \\
         \rowcolor{LightGreen} SymMIM$^{\dagger}$ & ViT-B/16 & 800 & \textbf{50.8} \\
         MAE$^{\dagger}$ \cite{he2022masked} & ViT-B/16 & 1600 & 48.0 \\
         \midrule
         \textcolor{gray}{scratch} & \textcolor{gray}{ViT-L/16} & \textcolor{gray}{-} & \textcolor{gray}{49.9} \\
         MoCo-v3 \cite{chenempirical} & ViT-L/16 & 600 & 49.1 \\
         BEiT$^{\dagger}$ \cite{bao2021beit} & ViT-L/16 & 800 & 53.3 \\
         ConMIM$^{\dagger}$ \cite{yi2022masked} & ViT-L/16 & 800 & 53.7 \\
         \rowcolor{LightGreen} SymMIM$^{\dagger}$ & ViT-L/16 & 800 & \textbf{54.1} \\
         MAE \cite{he2022masked} & ViT-L/16 & 1600 & 53.6 \\
    \bottomrule
    \end{tabular}
\end{table}

\section{Ablations}
\subsection{Analysis of learning objective}
In this section, we experiment with each additional learning objective in the total loss for the ablation study.
We progressively test with different sets of loss terms from \ref{eq:mimcon} and report the resulting accuracy on ImageNet-1K datasets.
The results, detailed in \ref{tab:objective} show an optimum when using the final objective in \ref{eq:mimcon}.
Interestingly, we also see that contrastive objective in \ref{eq:con} contributes the most toward the overall performance.

\begin{table}[!ht]
    \caption{Ablation of our multiple learning objectives.
    To validate our multiple learning objectives design, we report the results of our method with different sets of loss terms.
    Numbers are reported using a ViT-S/16 backbone.}
    \label{tab:objective}
    \centering
    \begin{tabular}{@{}c|c|c|c@{}}
    \toprule
         $\mathcal{L}_{\text{rec}_1}$ & $\mathcal{L}_{\text{rec}_2}$ & $\mathcal{L}_{\text{con}}$ & Acc. \\
    \midrule
        \ding{52} & \ding{56} & \ding{56} & 81.7 \\
        \ding{52} & \ding{52} & \ding{56} & 81.9 \\
        \ding{52} & \ding{56} & \ding{52} & 82.7 \\
        \ding{52} & \ding{52} & \ding{52} & \textbf{83.0} \\
    \bottomrule
    \end{tabular}
\end{table}

\subsection{Masking ratio probing}
\begin{figure}[!ht]
  \centering
  \includegraphics[width=\linewidth]{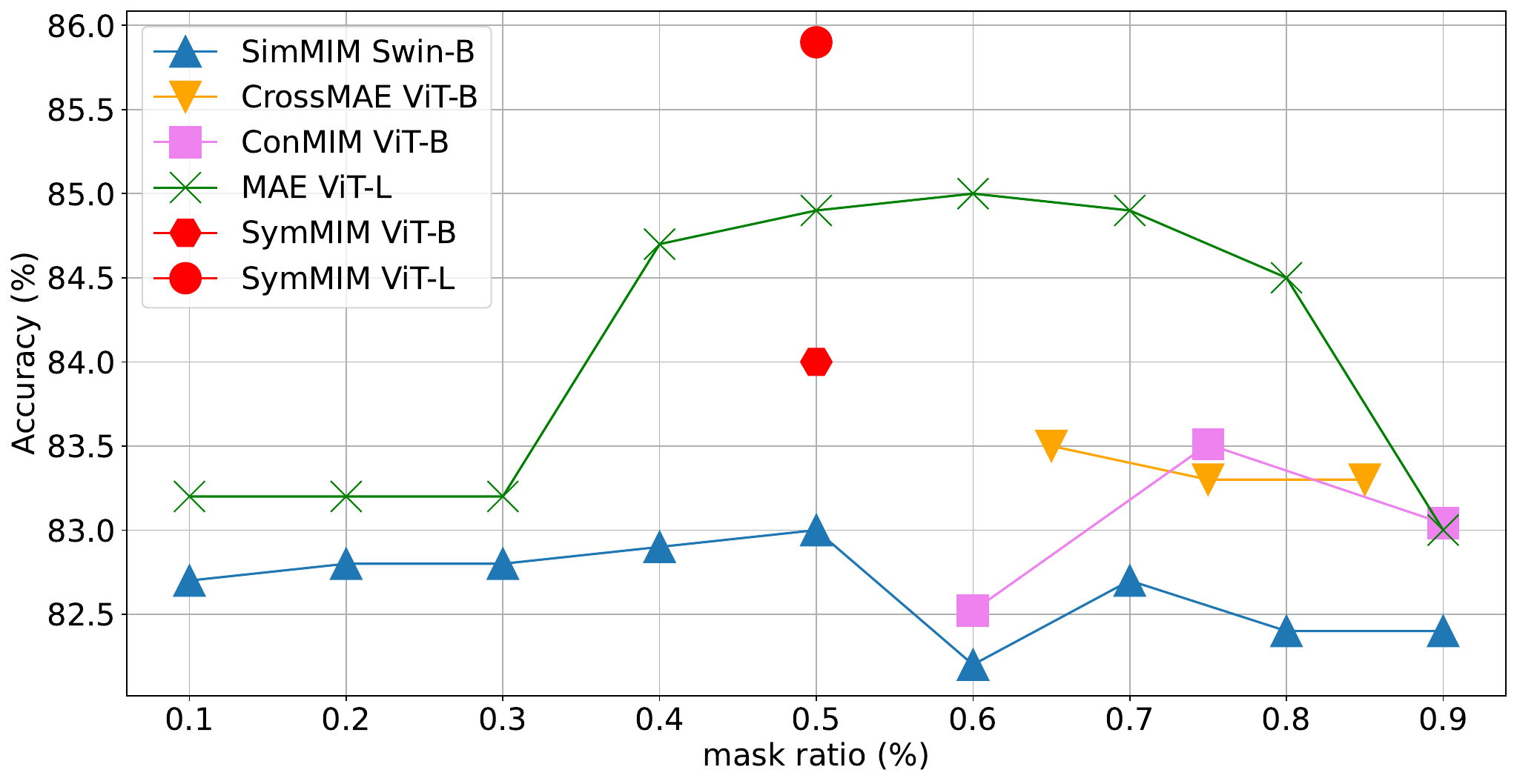}
  \caption{Masking ratio comparison.
  Traditional MIM such as SimMIM \cite{xie2022simmim} and MAE \cite{he2022masked} need to performs the masking ratio probing to find the optimal ratio, this results in a very expensive procedure.}
  \label{fig:masking-ratio}
\end{figure}

In this section, we conduct the ablation study on the masking ratio choices.
Figure \ref{fig:masking-ratio} shows the influence of the masking ratio towards the fine-tuning performance.
The accuracy fluctuates and is not stable for different methods and ratios.
This behavior shows that for different MIM methods and different datasets, we need to perform the probing again to fine the optimal ratio.
In contrast to that, our proposed method, SymMIM, using only one ratio of 50\% for both small and large masks.
SymMIM eliminates the probing procedure, which costly and inefficient approach while achieving SOTA across benchmarks.

\subsection{Visualization}
In this section, we delve into comprehending the proposed approach and critical designs through visualizations.
All the example images are sourced from the ImageNet-1K validation set.

\begin{figure}[!ht]
  \centering
  \begin{subfigure}{\linewidth}
    \includegraphics[width=\linewidth]{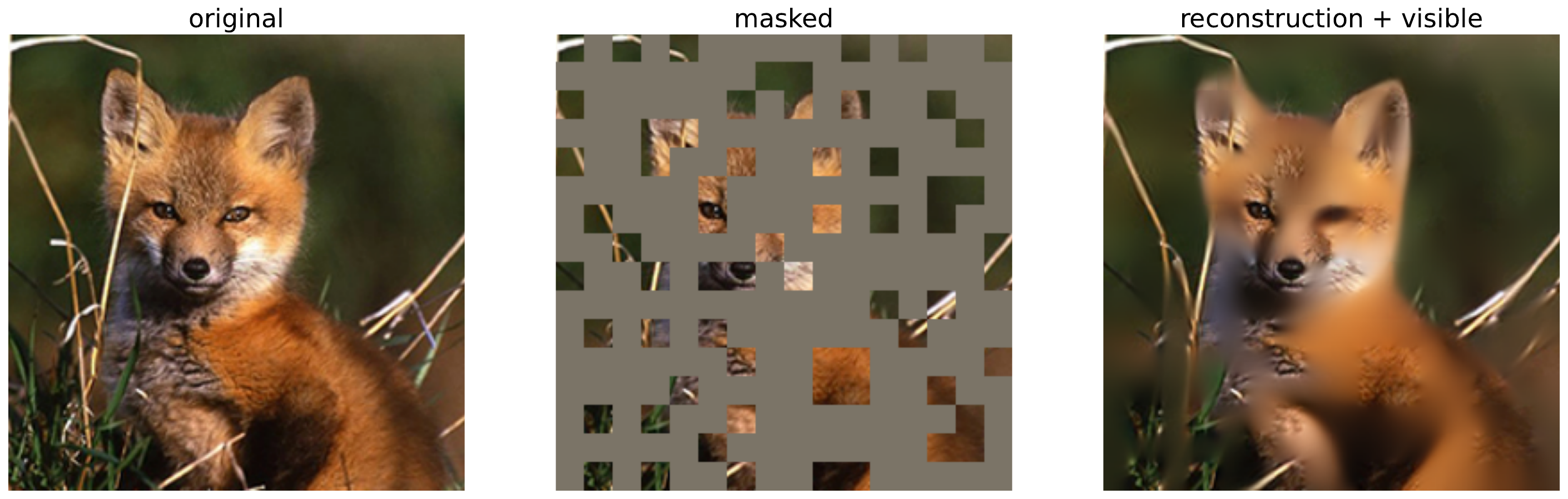}
    \caption{Random masking}
    \label{fig:random}
  \end{subfigure}
  \begin{subfigure}{\linewidth}
    \includegraphics[width=\linewidth]{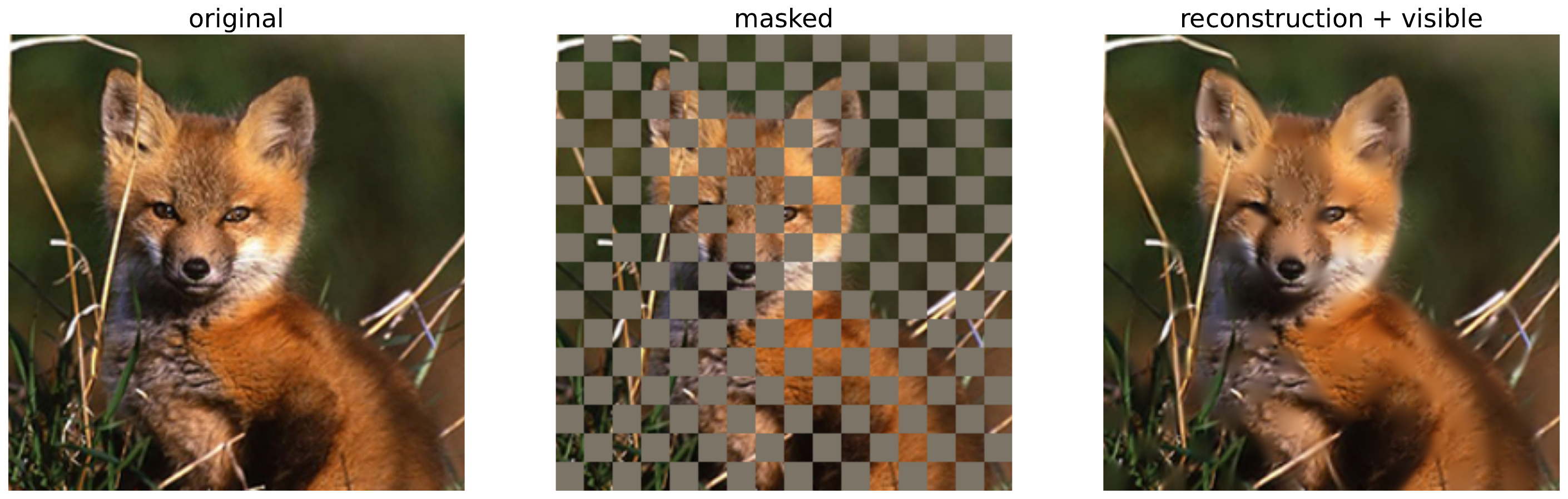}
    \caption{Very small size symmetric masking}
    \label{fig:verysmall}
  \end{subfigure}
  \begin{subfigure}{\linewidth}
    \includegraphics[width=\linewidth]{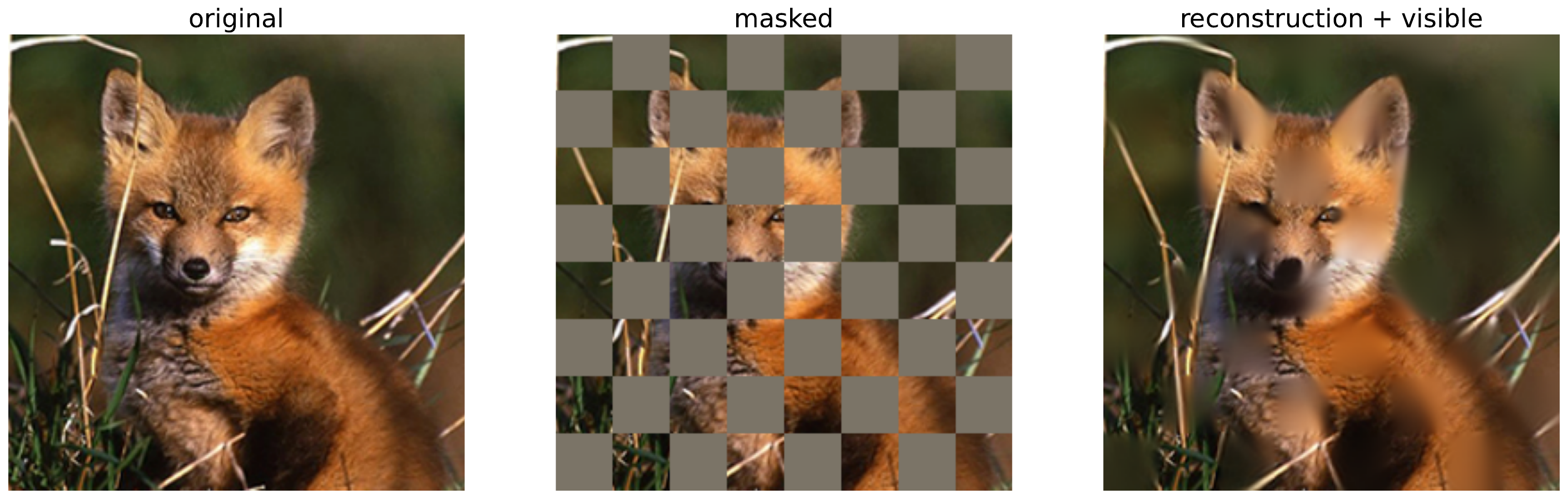}
    \caption{Small size symmetric masking}
    \label{fig:small}
  \end{subfigure}
  \begin{subfigure}{\linewidth}
    \includegraphics[width=\linewidth]{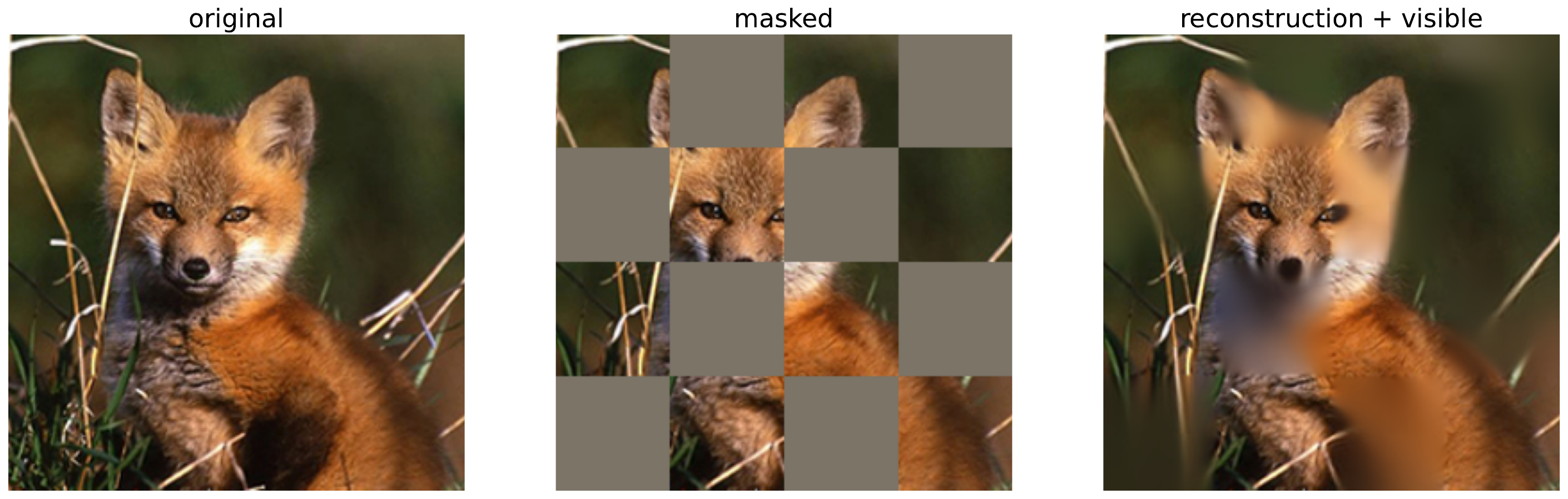}
    \caption{Large size symmetric masking}
    \label{fig:large}
  \end{subfigure}
  \caption{Recovered images using four different mask types: random masking, very small size symmetric masking, small size symmetric masking, and large size symmetric masking.}
  \label{fig:masking-strategy}
\end{figure}

\ref{fig:masking-strategy} illustrates the recovered images using various human-designed masks.
These masks serve to reveal the capabilities learned through masked image modeling.
Specifically, the human-designed masks include a random mask, a very small size mask, a small size mask, and a large size symmetric mask.

As we can see, the recovered image using a large size symmetric mask preserves much more global information, while the small and very small size masks are better at recovering local information.
Therefore, the model trained with our proposed method, which utilizes both large and small size masks is more robust.

\section{Conclusion}
In this study, we introduce a straightforward yet impactful symmetric masking strategy that simplifies hyperparameter tuning and masking ratio exploration.
We then present SymMIM, a novel masked image modeling framework that leverages this symmetric masking approach.
SymMIM combines reconstruction loss and contrastive loss to enhance representation learning effectively.
Despite its simplicity, SymMIM achieves remarkable performance across various benchmarks, setting a new state-of-the-art record.
Notably, our proposed SymMIM significantly accelerates the training process—requiring only a single pre-training step—while other methods necessitate linear probing nine to ten times.
\clearpage

\bibliographystyle{splncs04}
\bibliography{egbib}
\end{document}